%% file: main.tex
\documentclass[10pt,twocolumn,letterpaper]{article}

\usepackage{cvpr}              
\usepackage{multirow}
\usepackage{graphicx}


\definecolor{cvprblue}{rgb}{0.21,0.49,0.74}
\usepackage[pagebackref,breaklinks,colorlinks,allcolors=cvprblue]{hyperref}

\def\paperID{9935} 
\def\confName{CVPR}
\def\confYear{2025}

\title{MM-OR: A Large Multimodal Operating Room Dataset \\for Semantic Understanding of High-Intensity Surgical Environments}

\author{
Ege Özsoy$^{1,2}$ \; Chantal Pellegrini$^{1,2}$ \; Tobias Czempiel$^{1}$ \; Felix Tristram$^{1,2}$ \; Kun Yuan$^{1,2}$ \\
David Bani-Harouni$^{1,2}$ \; Ulrich Eck$^{1}$ \; Benjamin Busam$^{1,2}$ \; Matthias Keicher$^{1,2}$ \; Nassir Navab$^{1,2}$ \\[0.5em]
$^{1}$ Technical University of Munich
\quad $^{2}$ Munich Center for Machine Learning\\
{\tt\small ege.oezsoy@tum.de}
}

\begin{document}

\twocolumn[{
\maketitle
    \centering
    \includegraphics[width=1.0\textwidth]{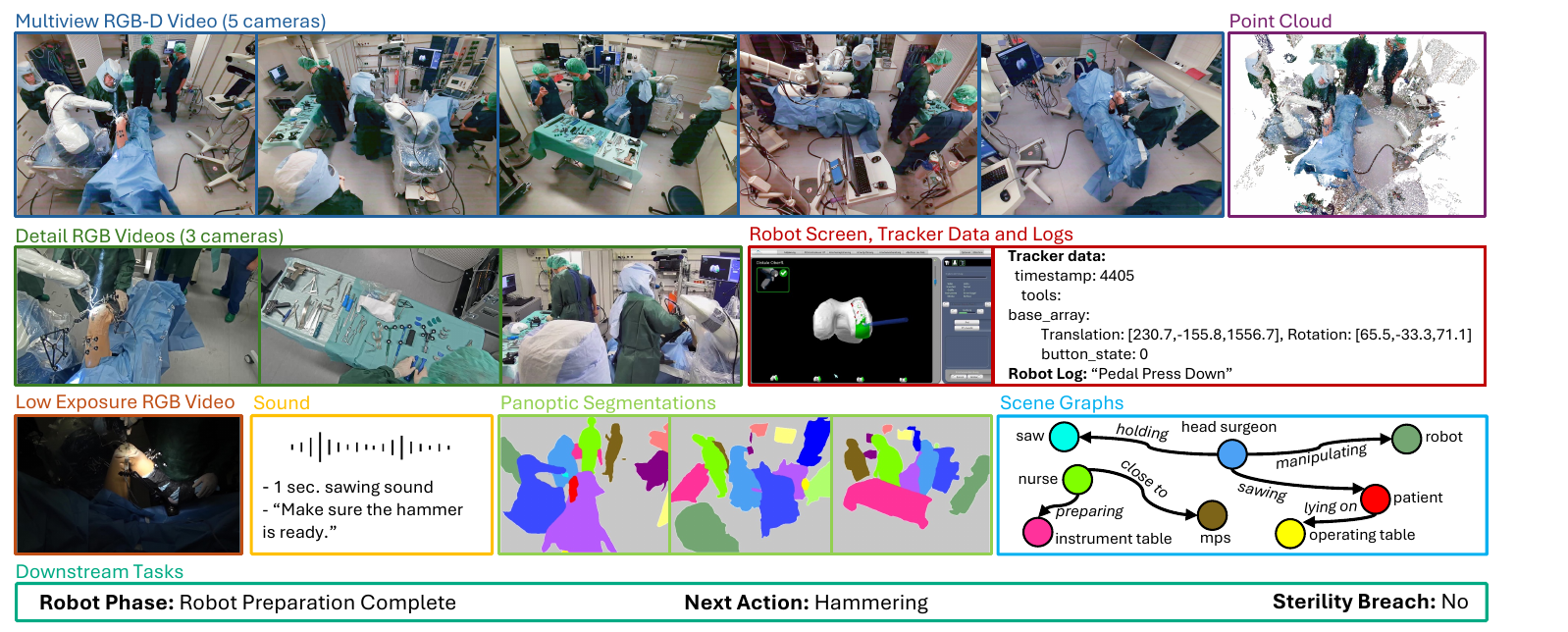} 
    \captionof{figure}{Overview of a single timepoint in MM-OR, illustrating the multimodal data provided for each sample: RGB-D video from multiple angles, detailed RGB views, low-exposure video, point cloud data, robot screen and tracker logs, audio and speech transcripts, panoptic segmentations, semantic scene graphs, and downstream task annotations such as robot phase, next action, and sterility breach status.\looseness=-1} 
    \label{fig:dataset_overview}
    \bigskip
}]

\input{sec/0_abstract}    
\input{sec/1_intro}
\input{sec/2_related_work}
\input{sec/3_dataset}
\input{sec/4_method}
\input{sec/5_experiments}
\input{sec/6_conclusion}
{
    \small
    \bibliographystyle{ieeenat_fullname}
    \bibliography{main}
}

\input{sec/X_suppl}

\end{document}

%% file: sec/0_abstract.tex
\begin{abstract}
Operating rooms (ORs) are complex, high-stakes environments requiring precise understanding of interactions among medical staff, tools, and equipment for enhancing surgical assistance, situational awareness, and patient safety. Current datasets fall short in scale, realism and do not capture the multimodal nature of OR scenes, limiting progress in OR modeling. To this end, we introduce MM-OR, a realistic and large-scale multimodal spatiotemporal OR dataset, and the first dataset to enable multimodal scene graph generation. MM-OR captures comprehensive OR scenes containing RGB-D data, detail views, audio, speech transcripts, robotic logs, and tracking data and is annotated with panoptic segmentations, semantic scene graphs, and downstream task labels. Further, we propose MM2SG, the first multimodal large vision-language model for scene graph generation, and through extensive experiments, demonstrate its ability to effectively leverage multimodal inputs. Together, MM-OR and MM2SG establish a new benchmark for holistic OR understanding, and open the path towards multimodal scene analysis in complex, high-stakes environments. Our code, and data is available at \url{https://github.com/egeozsoy/MM-OR}.

\end{abstract}

%% file: sec/1_intro.tex
\section{Introduction}
\label{sec:intro}

\begin{figure}[ht]
    \centering
    \includegraphics[width=1.0\linewidth]{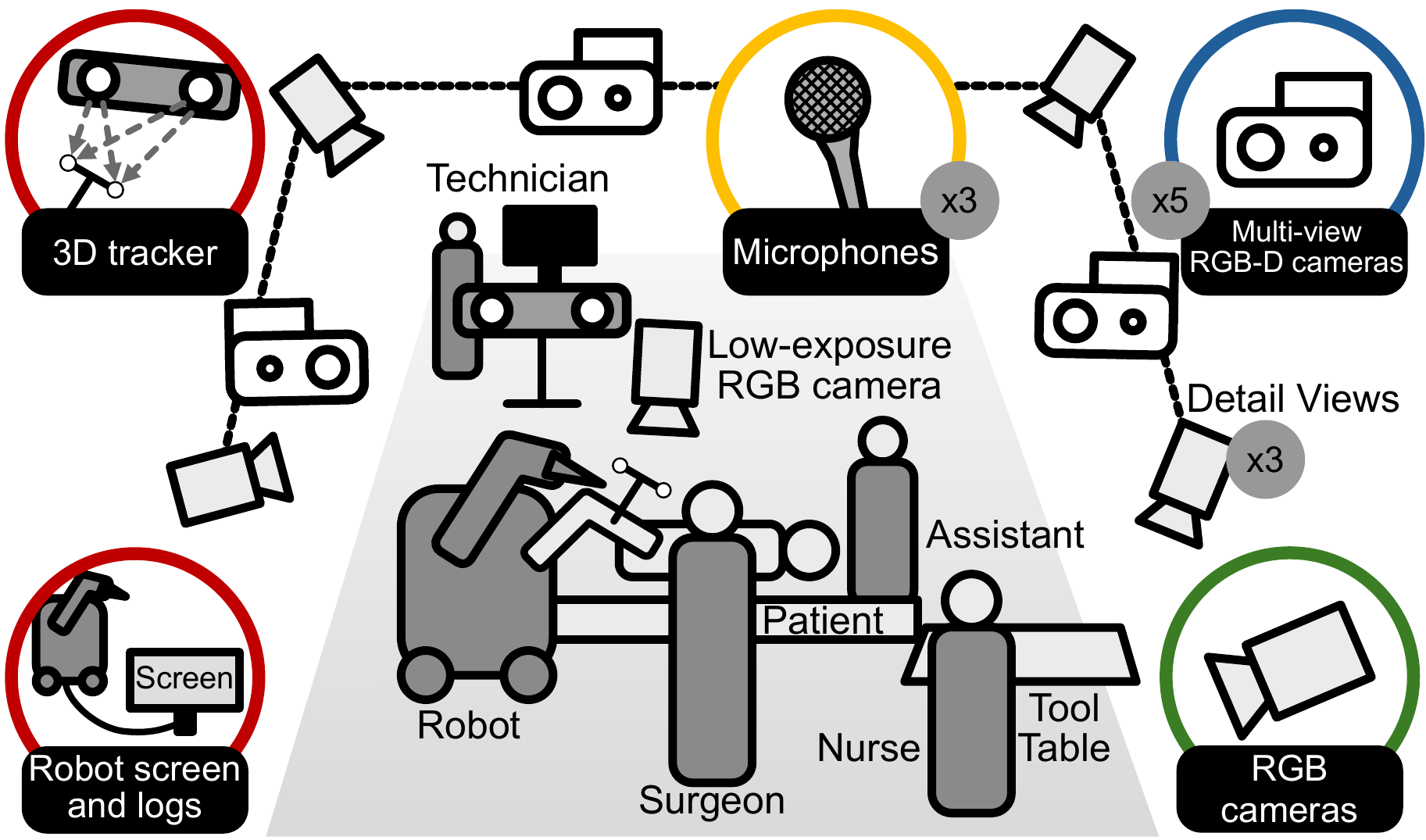}
    \caption{Recording setup and sensors overview. A grey circle by each sensor shows quantity; if absent, the sensor count is one.}
    \label{fig:overview}
\end{figure}

The operating room (OR) is one of the most complex and high-stakes environments needing accurate scene understanding~\cite{lalys2014,maier-hein_sds_2017}. However, the OR domain poses unique challenges due to the lack of public operating room datasets, capturing the distinct visual patterns, specialized surgical tools and equipment, environment, dynamic surgical actions, and complex interactions among clinical staff. 

Modern computer vision has been propelled by foundational datasets, such as MNIST~\cite{deng2012mnist}, CIFAR-10~\cite{krizhevsky2009learning}, and ImageNet~\cite{imagenet}. In specialized domains, such as autonomous driving, data efforts like KITTI~\cite{geiger2013vision}, Cityscapes~\cite{cordts2016cityscapes}, and Waymo Open Dataset~\cite{sun2020scalability} have similarly driven progress in scene understanding of dynamic settings. 
Likewise, the surgical field is a domain with immense room and potential for computer vision. However, acquiring data at scale in the OR is a challenge due to access, privacy restrictions, and acquisition hardware that must not interfere with the surgical workflow. Therefore, the few existing datasets are either proprietary~\cite{sharghi2020automatic} or lack in scale~\cite{ozsoy20224d, mvor}.

The surgical data science (SDS) field has made considerable strides in recent years, with a growing body of work dedicated to analyzing specific components of surgical procedures such as classifying surgical actions~\cite{tripnet, rendezvous}, identifying surgical phases~\cite{czempiel2021opera}, detecting instrument usage~\cite{jin2018tool}, estimating hand or human poses~\cite{wang2023pov,srivastav2018mvor} and anonymization~\cite{bastian2023disguisor}. The unique demands of SDS in complex, high-stakes environments like surgery introduce challenges and opportunities for advancing CV techniques in highly dynamic and nuanced contexts~\cite{wu2025self,hu2025ophnet,liu2023skit,yuan2024procedure,wu2024surgicai,montana2024saramis,zhou2023text}. Recently, the focus has shifted towards holistic OR domain modeling~\cite{ozsoy2023holistic}, aiming to capture the complete surgical scene and its multifaceted interactions. Özsoy et al.~\cite{ozsoy20224d} introduced the 4D-OR dataset, which is to this date the biggest public operating room dataset and the only one containing scene graph annotations. These annotations facilitate scene graph generation~\cite{image_retrieval_using_scene_graphs} as the main task, which enables the modeling of the interactions between the entities in the surgery, such as the head surgeon performing a drilling procedure on the patient's knee with assistance from the assistant surgeon. While 4D-OR has been instrumental in advancing OR modeling, its limitations restrict its utility for developing more advanced models. It is relatively small, comprising only 6,743 timepoints, constrained to a single day of recording with non-medical actors, and has little variability, occlusions, and complex interactions, limiting its realism. Additionally, 4D-OR lacks segmentation annotations and relies only on external room cameras, further limiting the dataset’s ability to capture intricate details and other modalities and signals within the OR. While 4D-OR has served as a critical stepping stone, the community needs a new dataset to address the limitations and continue advancing in surgical scene understanding.

To this end, we introduce MM-OR, a new public dataset designed to advance OR modeling through increased scale, realism, and multimodality, and the first dataset to enable multimodal scene graph generation. MM-OR consists of robotic total and partial knee replacement surgeries, integrating a comprehensive range of data sources, including multi-view RGB-D data, high-resolution RGB views, a low-exposure tracking camera, audio, speech transcripts, real-time robotic system logs, screen recordings of the robot interface, and infrared tracking. This multimodal setup enables MM-OR to capture a detailed and dynamic representation of the OR environment, accommodating the complex interactions and contextual variations inherent in surgical procedures. Beyond multimodality, MM-OR emphasizes realism. While recorded in a simulated setting, the procedures are performed by real surgeons on an anatomically accurate model, closely mirroring the clinical environment. In terms of scale, MM-OR surpasses prior datasets by a substantial margin. With about 15 times more raw timepoints and four times as many annotations, MM-OR captures a wider range of surgical activities, interactions, and scenarios, providing a more extensive and varied dataset for model training. Finally, MM-OR sets a new benchmark by being the largest publicly available OR dataset and the first to include panoptic segmentations, significantly enhancing the spatial and contextual modeling of surgical environments.

Building upon MM-OR, we introduce MM2SG, the first multimodal large language model (MLLM) for scene graph generation. MM2SG is capable of processing diverse data modalities end-to-end to generate comprehensive scene graphs, realized by using a multimodal encoding module that encodes varied inputs — including audio cues, point cloud data, speech, robotic system logs, tracking data, and segmentation masks into token embeddings. These embeddings are then jointly fine-tuned with a large language model (LLM) for a more holistic scene understanding. We conduct extensive modality ablation studies to assess the impact of individual inputs, demonstrating how each modality can contribute to scene graph generation. In summary, our main contributions are:
\begin{enumerate}
    \item \textbf{MM-OR Dataset:} MM-OR is the first dataset to enable multimodal scene graph generation, and the largest operating room (OR) dataset, including diverse modalities, supporting comprehensive OR scene modeling.
    \item \textbf{MM2SG Model:} We present MM2SG, the first multimodal large vision-language model (MLLM) for scene graph generation, which effectively leverages diverse data inputs for holistic OR scene understanding.
\end{enumerate}

\noindent
We believe this work not only advances OR modeling but also sets a foundation for multimodal scene analysis in other high-stakes, dynamic environments beyond surgery.

%% file: sec/2_related_work.tex
\section{Related Work}
\label{sec:related_work}

\noindent{\textbf{Surgical Data Science.}}
In the field of surgical data science (SDS), understanding surgical scenes and activities is essential for various applications, such as automated skill assessment~\citep{jin2018tool,liu2021towards,goodman2024analyzing}, decision support~\citep{yuan2021surgical,twinanda2016endonet,maier2017surgical}, and workflow efficiency improvement~\citep{yeung2019computer}. Prior works employ deep learning models to identify anatomies~\citep{murali2023latent}, instruments~\citep{jin2018tool}, and their interactions~\citep{nwoye2020recognition}. More recently, large-scale internal datasets have been curated, allowing the development of foundation models, and enhancing models' adaptability across a broader range of surgical scenarios~\citep{wang2023foundation,hu2025ophnet,yuan2023learning,yuan2024procedure}. However, all these works focus only on internal patient views, such as laparoscopy or microscopy. 

A few works focus on external surgery views with the existing datasets being either non-temporal and limited in size~\citep{srivastav2018mvor} or proprietary~\cite{sharghi2020automatic} and not including modalities beyond video streams~\citep{srivastav2018mvor,sharghi2020automatic,ozsoy20224d}. Multiple works proposed surgical scene graph generation approaches based on RGB-D data~\cite{ozsoy2023holistic}, introducing memory scene graphs~\cite{ozsoy2023} and leveraging large vision-language models for knowledge integration~\cite{ozsoy2024oracle}. However, all these datasets and methods rely only on video data, limiting the comprehensiveness of scene understanding by excluding information sources, such as machine signals, sound, and other unique intra-operative modalities. Our proposed dataset and model address this gap by incorporating additional modalities, providing a more versatile and comprehensive approach for generalized surgical scene understanding.

\noindent{\textbf{Scene Graph Generation and Multimodal Scene Understanding.}} Scene graph generation (SGG)~\cite{image_retrieval_using_scene_graphs} structures visual scenes into graphs, representing entities and their relationships. Existing scene graph datasets span from Visual Genome~\cite{visual_genome} and PSG~\cite{yang2022panoptic} for images, to video~\cite{yang2023panoptic} and 4D data~\cite{yang20244d, ozsoy20224d}. In parallel, large scene understanding datasets~\cite{grauman2022ego4d, li2024mvbench, miech2019howto100m, liu2022hoi4d} focusing on tasks such as video captioning, QA, and language-guided video analysis, have pushed scene comprehension further by combining text, multi-view video, and audio. EgoExo4D~\cite{grauman2024ego} broadens multimodal scene understanding with a multi-view egocentric setup, capturing human-object interactions, eye-gaze, audio, and a static SLAM point cloud. While these datasets have contributed significantly to scene and multimodal understanding, they face limitations related either to modality diversity, lack of scene graph annotations, or temporal length, especially in the context of high-stakes, procedural domains like operating rooms. MM-OR is the first multimodal scene graph dataset and provides hour-long recordings of complex tasks that require human-robot interaction.

\noindent{\textbf{Multimodal Large Language Models.}}
Multimodal Large Language Models (MLLMs) enhance LLMs by incorporating additional modalities with text-only LLMs, effectively merging the embedded knowledge and text comprehension of LLMs with multimodal understanding. Initial works focused on image integration~\cite{alayrac2022flamingo,li2023blip,liu2023visualllava} and more recently video understanding~\cite{li2023videochat,lin-etal-2024-video,ozsoy2024oracle}. Some recent MLLMs incorporate additional modalities, such as depth information, audio, inertial measurements, or MRI data ~\cite{shukor2023unival,han2024onellm,moon2024anymal} and show promise for multi-modal understanding using MLLMs for tasks such as visual grounding, captioning, question answering, reasoning, or recommendations. Nevertheless, until now there exists no public multimodal dataset or MLLM for holistic OR understanding.

%% file: sec/3_dataset.tex
\section{MM-OR Dataset}
\label{sec:dataset}

The MM-OR dataset addresses the critical need for a large-scale, multimodal, and realistic dataset to support advanced OR modeling. While previous datasets such as 4D-OR~\cite{ozsoy20224d} have established foundational insights, their limitations in scale, realism, and data variety restrict the development of models. MM-OR overcomes these constraints by capturing 17 full-length ($\sim90$ minutes) and 22 short clips of robotic total and partial knee replacement surgeries, one of the most common procedures in orthopedic surgery, with tens of thousands performed annually worldwide. These surgeries also share essential workflows and interactions with many other procedures, making MM-OR broadly applicable. Recorded over multiple days with diverse clinical staff, including varying head surgeons and team compositions, each surgical sequence mirrors real-world procedures, presenting varied and challenging conditions, including frequent occlusions, lighting changes, and dynamic interactions among team members and equipment. To closely replicate a real surgical experience, we used a plastic medical-grade knee model encased in foam to simulate the structure of the leg. This setup allowed the surgeon to follow typical steps, such as scalpel access and bone sawing. Further details about the phantom are provided in the supplementary materials. MM-OR’s diverse data modalities, extensive annotations with panoptic segmentations, scene graphs, and downstream task labels, and realistic clinical settings mark a significant step forward. With 92,983 total timepoints and 25,277 annotations totaling 500GB of data, MM-OR sets a new benchmark for comprehensive surgical scene understanding and OR modeling.

\subsection{Data Modalities}
In the OR, non-visual data such as audio cues, voices or tool signals offer critical insights, highlighting the importance of a multimodal approach that goes beyond vision alone, particularly when additional hardware like robots are involved. Each modality was synchronized at the hardware level to ensure precise temporal alignment, and a movie clapper was used for manual calibration across modalities. MM-OR captures a rich, multimodal representation of the OR, including the following data sources:

\noindent\textbf{Multi-view RGB-D Video Stream from Room Cameras.} Five  Kinect RGB-D cameras on the ceiling capture a synchronized 3D view of the surgical scene. Alongside time-synchronized RGB images, MM-OR includes depth maps, and camera parameters, facilitating the creation of colored point clouds for detailed spatial modeling.

\noindent\textbf{RGB-Only Detail Views.} Complementing the room cameras, we include three higher dynamic range cameras focused on key areas of the OR, such as the surgical field and instrument table. These enable precise identification of instruments, hand movements, and subtle procedural cues.

\noindent\textbf{Low-Exposure RGB Camera.} Bright OR lighting often causes overexposure in standard digital cameras, making it difficult to capture clear images of the surgical field. To address this, we utilize a low-exposure RGB camera from the robotic surgery tracking setup. This camera, positioned for unobstructed views of the surgical field to maintain tracking integrity in surgery, ensures that even under intense lighting, tool trajectories and tool usage is clearly captured.

\noindent\textbf{Audio and Speech Transcripts.} Audio data from three wireless microphones worn by surgical staff, recording both tool sounds and verbal exchanges such as procedural intent, assistance requests, or team coordination, provide valuable insights into surgical workflow, situational awareness, and procedural flow that extend beyond visual data alone. For instance, actions like "hammering" could be detected more precisely through audio cues than through external visual views. To extract speech transcripts, the audio is transcribed using Whisper v3~\cite{radford2023robust} into timestamped text files.

\noindent\textbf{Robotic System Logs and Screen Recordings.} MM-OR extensively documents the robot’s operation with screen recordings of its user interface, the main display for the robot technician (MPS) and staff, and internal logs detailing its state and perception. These sources aid in identifying interactions like calibration and provide insights into the status and success of phases and nuanced tasks, such as knee bone sawing.

\noindent\textbf{3D Infrared Tracking.} The robotic knee replacement surgery leverages infrared marker-based tracking to enable spatial localization of the patient's knee, certain tools, and the robot. This data includes tracking information for the robot’s base array, the calibration array, two trackers attached to the patient’s knee, and specialized calibration tools used for registering specific bone regions. It provides reliable spatial information and context for interpreting surgical phases. For example, the presence and movement of the calibration tools can indicate the setup phase, while specific positioning of the knee trackers can inform models about alignment and registration stages.

Together, these multimodal data sources provide a comprehensive view of the OR, capturing both spatial and procedural dynamics essential for surgical scene understanding. This diversity enables MM-OR to support various tasks, from precise tool tracking to nuanced analysis of human-robot interactions within the surgical workflow.

\begin{figure*}
    \centering
    \includegraphics[width=\linewidth]{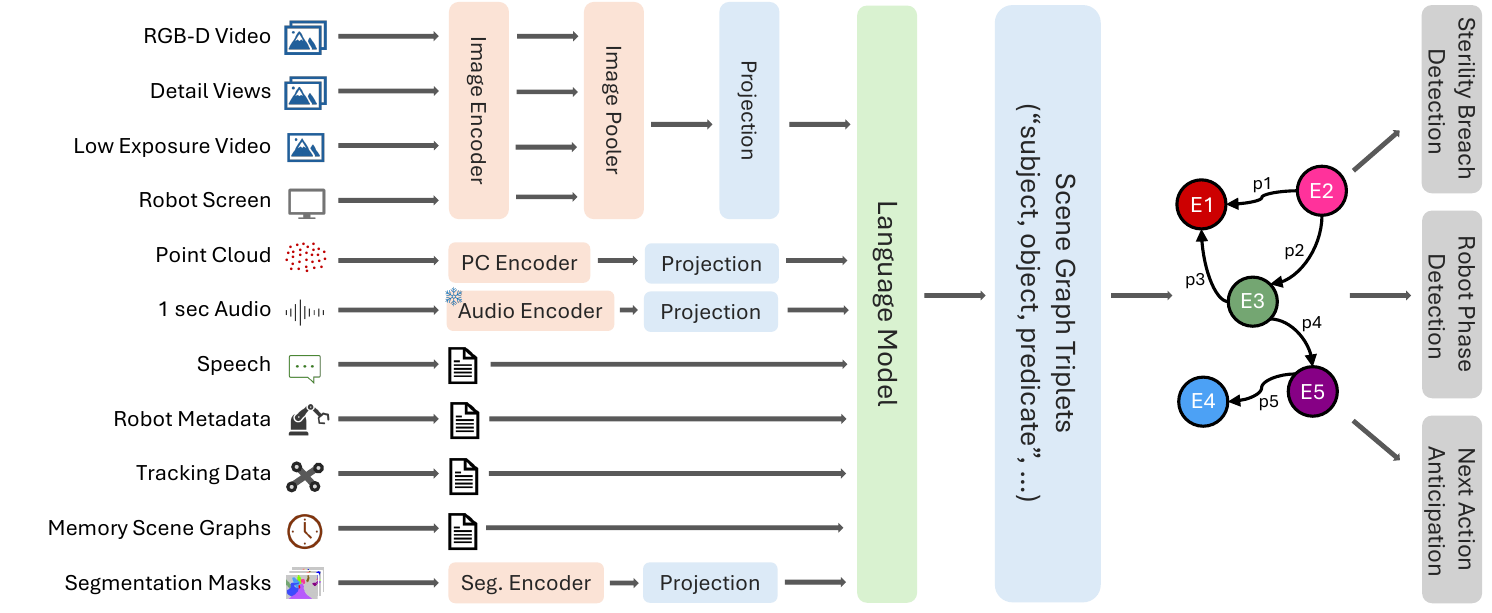}
    \caption{Overview of the proposed MM2SG architecture for multimodal scene graph generation. MM2SG processes a variety of data sources through specialized encoders, projecting them into a shared space. The language model generates scene graph triplets describing SGs with entities $E_i$ and predicates $p_i$. Downstream tasks leverage entire sequences of scene graphs rather than individual ones.}
    \label{fig:method_overview}
\end{figure*}

\subsection{Annotations}
The annotations in MM-OR support tasks that require a deep analysis of surgical workflows. Each annotation was performed by one annotator and reviewed by another.

\noindent\textbf{Panoptic Segmentation.} MM-OR is the first publicly available OR dataset to include panoptic segmentation, providing pixel-level annotations for surgical tools, staff, and equipment across three RGB-D camera views. These are created using a custom annotation tool that uses Segment Anything~\cite{kirillov2023sam} and AOT~\cite{yang2021aot}. First, the annotators outline objects of interest in an initial frame using a multi-click approach. The tracker then propagates segmentation masks to subsequent frames, which annotators review and correct, adding new objects if needed. Annotators segmented every frame during high-action scenes and every fifth frame during low-action periods, resulting in a mixture of manual and interpolated annotations. These can also be projected into 3D space, enabling accurate spatial OR modeling.

\noindent\textbf{Semantic Scene Graphs.} For every timepoint with manually labeled segmentations, MM-OR also includes manually annotated semantic scene graphs similar to 4D-OR~\cite{ozsoy20224d}, capturing the interactions between surgical staff, tools, equipment, and the robotic system. These annotations facilitate complex tasks such as activity recognition, situational awareness modeling, and human-robot interaction analysis, providing foundational data for understanding how entities interact dynamically in the OR. The entity and predicate counts, covering a wide range of surgical roles, tools, equipment, and actions, are summarized in the supplementary.

\noindent\textbf{Downstream Tasks.} To further extend MM-OR’s utility in surgical procedure understanding, we provide annotations for three clinically relevant downstream tasks:

\textbf{Robot Phase Prediction} labels key phases in the robot setup, like calibration or base array and saw installation. These annotations, provided at each timepoint, correspond to standard steps that staff follows to prepare the robot.

\textbf{Next Action Anticipation} in surgery is crucial for preventing accidents and ensuring a smooth transition between surgical steps. For instance, a robot could start preparing tools before the next surgical action. MM-OR includes annotations specifying when and what the next significant action in the surgery will be, enabling models that can predict upcoming events and systems that preemptively prepare the necessary resources or interventions.

\textbf{Sterility Breach Detection} In the OR, maintaining sterile fields is vital, and sterile and non-sterile staff and objects need to avoid cross-contact. MM-OR provides annotations for any breaches of sterility, indicating when inadvertent interactions occur. This task is of high clinical relevance, as it aims to minimize infection risks by identifying and flagging inadvertent contact between sterile and non-sterile entities.

\vspace{1em}

In addition to the full-length recordings, MM-OR includes 20 short clips capturing one to five-minute sequences where tasks are performed correctly or incorrectly. These clips highlight specific actions or procedural phases, enhancing MM-OR’s utility for tasks such as anomaly detection, and error identification. Beyond the OR, its extensive temporal and interaction data offer applications in broader computer vision tasks, including panoptic scene graph generation~\cite{yang2022panoptic}, multimodal scene understanding~\cite{grauman2024ego}, video segmentation~\cite{kim2020video}, long-form video understanding~\cite{tan2024koala,he2024ma}, multi-agent interaction~\cite{jeong2024multi}, and action recognition~\cite{chen2024align}.

%% file: sec/4_method.tex
\section{Methodology}
\label{sec:method}
We designed a novel approach, MM2SG, for generating scene graphs from multimodal data captured during surgical procedures. While previous methods focus on visual signals such as images or point clouds, MM2SG takes a holistic approach by integrating camera data with a diverse set of other data modalities such as audio and speech, real-time robotic system logs, screen recordings of the robotic interface, infrared tracking data, segmentation masks, and temporal context. We build upon ORacle\footnote{https://github.com/egeozsoy/ORacle}~\cite{ozsoy2024oracle} and model the scene graph generation as a language modeling task.

\subsection{Architecture Overview}
MM2SG has two main components: a multimodal processing module and a large language model (LLM). The multimodal processing module takes in a variety of input data and encodes each modality into token embeddings that can be processed by the LLM. The LLM then takes these multimodal token embeddings and autoregressively generates the scene graph as a list of triplets.

\begin{table*}[h]
\centering
\setlength{\tabcolsep}{3.3pt} 
\begin{tabular}{l c c c c c c c |c}
\hline
\textbf{Model} & \textbf{Room Cameras} & \textbf{Other Cams.} & \textbf{PC} & \textbf{Audio/Transcript} & \textbf{Robot Logs/Tracker} & \textbf{Seg} & \textbf{Temporality} & \textbf{F1} \\
\hline
PSG~\cite{yang2022panoptic} & \checkmark & & & & & & & 0.328 \\
ORacle~\cite{ozsoy2024oracle} & \checkmark & & & & & & & 0.472 \\
\midrule
MM2SG & \checkmark & \checkmark & & & & & & 0.484 \\
MM2SG & \checkmark & \checkmark & \checkmark & & & & & 0.494 \\
MM2SG & \checkmark & \checkmark & \checkmark & \checkmark & & & & 0.484 \\
MM2SG & \checkmark & \checkmark & \checkmark & \checkmark & \checkmark & & & 0.502 \\
MM2SG & \checkmark & \checkmark & \checkmark & \checkmark & \checkmark & Pred & & 0.520 \\
MM2SG$\star$ & \checkmark & \checkmark & \checkmark & \checkmark & \checkmark & Pred & \checkmark & \textbf{0.529} \\
\midrule
MM2SG & \checkmark & \checkmark & \checkmark & \checkmark & \checkmark & GT & & 0.519 \\
MM2SG & \checkmark & \checkmark & \checkmark & \checkmark & \checkmark & GT & \checkmark & \textbf{\textit{0.534}} \\
\hline
\end{tabular}
\caption{Scene graph generation results on MM-OR. Modalities are abbreviated as follows: 'Other Cams.' for detailed-view and low exposure tracking camera, 'PC' for Point Cloud, 'Seg' for segmentation data ('GT' for ground truth and 'Pred' for predicted masks). Macro F1 scores indicate that MM2SG outperforms the baseline PSG~\cite{yang2022panoptic}, a non-domain-specific model. Moreover, the progressive addition of modalities leads to improved performance, with the highest F1 score obtained when all modalities and temporality are used. Unless otherwise specified, references to MM2SG denote the $\star$ version.}
\label{tab:sgg_results}
\end{table*}

\subsubsection{Modality-Specific Encoding}
We use modality-specific encodings to capture each modality effectively:\\
\textbf{Images.} We process a variable number of images from the multi-view RGB-D cameras, additional high-resolution detail views, tracking camera, and screen recordings of the robotic interface together. Each image is first encoded using a CLIP vision encoder~\cite{radford2021learning}, and then the patch embeddings are concatenated and processed by a transformer-based image pooler, of which only the first N output embeddings are used as image tokens, summarizing the images to a fixed-length representation. The image encoder and image pooler are trained end-to-end with the LLM.\\
\textbf{Point Cloud.} We train a Point Transformer V3~\cite{wu2024point} end-to-end with the rest of the model to encode 3D point cloud data into a single token representation.\\
\textbf{Audio.} Audio data is encoded using a pre-trained and frozen CLAP audio encoder~\cite{elizalde2023clap}, again producing one token representation per audio clip. For every timepoint, we encode the audio from the last second.\\
\textbf{Speech.} We pre-process the audio recordings and extract speech transcripts using a pre-trained speech-to-text model~\cite{radford2023robust}. This allows us to directly input the last five spoken sentences as text into the LLM.\\
\textbf{Real-time Robot System Logs.} We extract high-level information about the robot's state, including the current phase and action, from the robotic logs and convert this to a textual representation, directly inputted to the LLM.\\
\textbf{Tracking Data.} For each timepoint, we identify the visible tools from the infrared tracker data and extract their translation and rotation. This information is then provided as structured text input in the prompt. \\
\textbf{Segmentation Masks.} Segmentation masks are encoded using a small end-to-end trained CNN, where each mask is converted into one token embedding.\\
\textbf{Temporality.} Following previous works~\cite{ozsoy2023,ozsoy2024oracle}, we incorporate temporal information describing the surgical procedure up to the current time point using memory scene graphs. For the most recent timepoints, we include full scene graph predictions as a list of triplets. The long-term context is summarized by only keeping each occurring triplet once in the order they were first predicted. Overall, the short-term information provides a detailed understanding of all recent actions, while the long-term context allows to estimate the approximate position within the surgery.

All modality-specific token embeddings are projected into the language model space by a linear projection layer, concatenated, and provided as input sequence to the LLM.

\subsubsection{Scene Graph Generation}
Using the multimodal features, MM2SG generates scene graphs as structured triplets in the form $(subject, object, predicate)$. The LLM autoregressively predicts these triplets, capturing both dynamic interactions, e.g. $(surgeon, drill, holding)$ and spatial relationships e.g. $(patient, lying\ on, operating\ table)$. The autoregressive prediction allows MM2SG to integrate contextual cues from all predicted triplets to guide subsequent predictions, resulting in a coherent representation of the OR scene.

\subsection{Multimodal Mix Augmentation}
To prevent overfitting and improve robustness, we apply a combination of modality dropping and mixing during training. These techniques encourage MM2SG to develop a more generalized understanding of each modality individually and in varied combinations, promoting more robust performance across diverse OR scenarios.

\noindent\textbf{Modality Dropping.} During training, each modality has a 50\% chance of being excluded from a sample, ensuring the model encounters and learns from incomplete and changing modality combinations.

\noindent\textbf{Multimodality Mixing.} We introduce a novel mixing augmentation to prevent the model from overfitting on specific combinations of modalities. During training, we identify samples with similar scene graphs to the current one and randomly replace certain modalities, such as audio, robotic metadata, tracker data, and speech transcripts, with those from these similar samples. This approach encourages MM2SG to recognize modality-specific patterns, like audio cues for drilling or hammering, without relying on fixed visual contexts, enhancing robustness.

\subsection{Downstream Tasks}
In addition to the primary task of scene graph generation, we offer straightforward baseline solutions for three downstream tasks, all using exclusively scene graphs as input:

\noindent\textbf{Robot Phase Prediction and Next Action Anticipation.} For these tasks, we fine-tune one LLM to take as input only the predicted scene graphs alongside the task name "phase recognition" or "next action". The LLM then outputs either the robot phase or the next anticipated action based on the current task. This scene graph-only approach is designed to provide baseline results and offers a lightweight setup.

\noindent\textbf{Sterility Breach Detection.} For sterility breach detection, we use a heuristic-based approach that directly analyzes predicted scene graphs to flag any interactions between sterile and non-sterile entities. This efficient and interpretable method provides a straightforward baseline for this clinically important task.

%% file: sec/5_experiments.tex
\section{Experiments}
\label{sec:experiments}
In this section, we outline the experimental setup, implementation details, and results for scene graph generation, our segmentation baselines, and downstream tasks. 

\begin{table}
    \centering
    \setlength{\tabcolsep}{5pt} 
    \begin{tabular}{lc|ccc}
        \toprule
        \textbf{Trained On} & \textbf{Temp.} & \textbf{4D-OR} & \textbf{MM-OR} \\
        \midrule
        \multirow{3}{*}{\textbf{MM-OR}} & \(\times\) & - & 64.6/59.6/59.3 \\
        & Online & - & 66.3/64.8/64.1 \\
        & Offline & - & 66.1/66.0/65.3 \\
        \midrule
        \multirow{3}{*}{\textbf{Both}} & \(\times\) & 64.9/67.8/67.5 & 64.1/62.3/61.7 \\
        & Online & 70.5/68.4/67.8 & 62.8/64.8/64.2 \\
        & Offline & \textbf{71.8}/\textbf{69.8}/\textbf{69.2} & \textbf{67.5}/\textbf{67.0}/\textbf{66.4} \\
        \bottomrule
    \end{tabular}
    \caption{Segmentation results on MM-OR and 4D-OR with different temporal variants and training datasets. Three scores are reporter per model, separated by "/", corresponding to VPQ with a window size 0, 4 and 8 respectively. Trained on "Both" implies both MM-OR and 4D-OR, whereas MM-OR stands for only MM-OR. Temporality is abbreviated as "Temp.", where "Online" uses only past context, and "Offline" the full context.}
    \label{tab:segmentation_results}
\end{table}

\begin{table}
    \centering
    \setlength{\tabcolsep}{10pt} 
    \begin{tabular}{lc|c}
        \toprule
        \textbf{Task} & \textbf{Method} & \textbf{F1} \\
        \midrule
        Sterility Breach Detection & Rule-based & 0.550 \\
        Next Action Anticipation & LLM-based & 0.354 \\
        Robot Phase Prediction & LLM-based & 0.569 \\
        \bottomrule
    \end{tabular}
    \caption{Initial baseline results on downstream tasks.}
    \label{tab:downstream_baseline_results}
\end{table}

\begin{figure*}
  \centering  
   \includegraphics[width=1.0\linewidth]{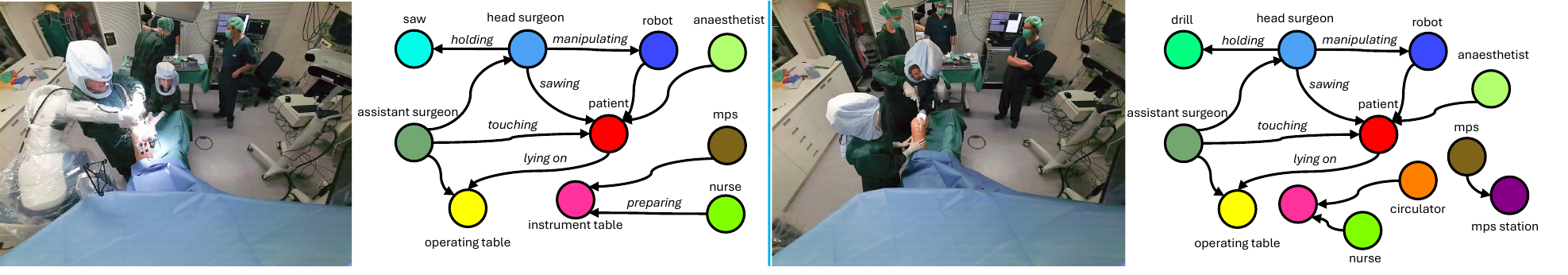}
   \caption{Qualitative examples from a test take in MM-OR, illustrating scene graph generation performance of MM2SG. Unlabeled edges indicate the "close to" predicate.}
   \label{fig:qual_examples}
\end{figure*}

\begin{table}[h]
    \centering
    \small
    \setlength{\tabcolsep}{5pt} 
    \begin{tabular}{c|c|c|c}
        \toprule
        & \textbf{Modality Dropping} & \textbf{Multimodality Mixing} & \textbf{F1} \\
        \midrule
        (1) & & & 0.436 \\
        (2) & \checkmark &  & 0.469 \\
        (3) & \checkmark & \checkmark & \textbf{0.520} \\
        \bottomrule
    \end{tabular}
    \caption{Ablation study on multimodality augmentation techniques. (1) None, (2) Modality dropping only, (3) Both modality dropping and multimodality mixing augmentations.}
    \label{tab:augmentation_ablation}
\end{table}

\subsection{Experimental Setup}
\noindent{\textbf{Multimodal Scene Graph Generation.}} In our primary experiments, we evaluate MM2SG against state-of-the-art models, including a top-performing panoptic scene graph generation model~\cite{yang2022panoptic} and the leading approach in surgical scene graph generation~\cite{ozsoy2024oracle}, each \textit{fine-tuned} on our dataset. To further assess the unique contributions of individual modalities, we incrementally add each one, starting with room camera views as the foundational input, to understand their impact on performance. Following prior work~\cite{ozsoy20224d}, we use macro F1-score over all predicates to evaluate scene graph generation. This metric gives equal weight to each predicate class, ensuring balanced evaluation across both common and rare relationship types.

\noindent{\textbf{Panoptic Segmentation.}} For segmentation baselines, we employ a state-of-the-art video segmentation model~\cite{zhang2023dvisplus}. Models are trained either on MM-OR alone or on a combined dataset of MM-OR and 4D-OR. To explore the impact of temporal data, we train three model configurations; only single frame, online (past context only), and offline (full temporal context). We measure the segmentation performance using Video Panoptic Quality~\cite{kim2020video}, with temporal window sizes of 0, 4, and 8 frames.

\noindent{\textbf{Downstream Task Baselines.}} For downstream tasks, we provide initial baseline results on three clinically relevant tasks: Robot Phase Prediction, Next Action Anticipation, and Sterility Breach Detection. Our approaches rely exclusively on scene graph data without additional multimodal inputs, and each task is assessed using macro F1 to ensure balanced performance across classes.

\subsection{Implementation Details} We initialize our MLLM with pretrained LLaVA-7B~\cite{liu2023visualllava} weights, utilizing Vicuna-7B~\cite{vicuna2023} as the LLM. Following ORacle~\cite{ozsoy2024oracle}, we employ temporal curriculum learning, beginning with non-temporal training and later fine-tuning with temporal data. For efficient fine-tuning, we use LoRA~\cite{hu2021lora} for the LLM, and also fine-tune the last 12 layers of the image encoder to adapt to the OR domain, with the number of image tokens being $N=576$. MM-OR dataset is split into nine train, three validation, and four test takes.  All models are trained for 20 epochs, with MM-OR comprising approximately 80\% of the training set, and 4D-OR the rest. We set the modality-dropping probability to 50\% per sample per modality. As the segmentation baselines, we use DVIS++~\cite{zhang2023dvisplus}, trained on either MM-OR alone or a combined dataset of MM-OR and 4D-OR~\cite{ozsoy20224d}. Since PSG~\cite{yang2022panoptic} processes a single image, we train it on each of the three views with segmentation masks. During inference, we fuse predictions from the three views by aggregating any triplet detected across views. All experiments are conducted on a single NVIDIA A40 GPU using PyTorch, with MM2SG training requiring approximately one week.

\subsection{Results}
\noindent\textbf{Scene Graph Generation.} In Table~\ref{tab:sgg_results}, we present the scene graph generation results of MM2SG under various modality configurations and compare it to the baseline models in panoptic~(PSG~\cite{yang2022panoptic}) and surgical scene graph generation~(ORacle~\cite{ozsoy2024oracle}). We find that OR-domain-specific models significantly outperform PSG, as their architecture is tailored to process OR-specific data, such as multiview input. Furthermore, we observe performance improvements when using more modalities, with the best result achieved when all modalities are used together, underscoring the value of a holistic, multimodal approach. Qualitative examples in Figure~\ref{fig:qual_examples} illustrate MM2SG’s ability to correctly capture challenging interactions, such as assisting and tool manipulations, even in cluttered scenes. However, challenges remain with infrequent relations, where performance is limited, likely due to the long-tailed distribution of relation types. This points to potential future work on addressing rare relationships in complex environments.

\noindent\textbf{Panoptic Video Segmentation.} In Table~\ref{tab:segmentation_results}, segmentation results across temporal configurations and datasets are shown. Temporal models, particularly when using full context, achieve the highest scores. This is most notable with the temporal window set to eight, emphasizing the advantage of long-range temporal data for accurately capturing dynamic surgical scenes. Additionally, training on both MM-OR and 4D-OR enhances segmentation performance compared to training only on MM-OR. This suggests improved generalization, with including 4D-OR samples helping the model adapt more robustly across surgical scenes.

\noindent\textbf{Downstream Tasks.}
Table~\ref{tab:downstream_baseline_results} presents macro F1 scores for Robot Phase Prediction, Next Action Anticipation, and Sterility Breach Detection. Using only scene graph data, our baselines achieve reasonable results across tasks, underscoring MM-OR's potential as a resource for downstream surgical tasks and enforce scene graphs as a lightweight and interpretable scene representation.

\noindent\textbf{Ablations on Augmentation Techniques.}  To assess the impact of multimodality augmentation strategies on model performance, we perform an ablation study on the MM2SG model using three configurations: (1) none, (2) modality dropping only, and (3) both modality dropping and multimodality mixing augmentations. Table~\ref{tab:augmentation_ablation} shows that combining both techniques yields the best performance.

\subsection{Limitations}
MM-OR advances OR scene modeling with its scale, realism, and multimodality, but it has limitations. It includes only total and partial knee replacements in one OR, potentially limiting generalization to other procedures or settings. The comprehensive acquisition setup, while effective, is costly and complex to scale across diverse ORs or real-time hospital use. Future work could enhance MM-OR’s scope with adaptable setups and broader surgical variety.

%% file: sec/6_conclusion.tex
\section{Conclusion}
We present MM-OR, the first large-scale, realistic multimodal dataset for OR modeling, enabling multimodal scene graph generation with diverse data sources including RGB-D, high-resolution views, audio, speech, robotic logs, and tracking data. Similar to how large datasets have driven progress in fields like autonomous driving and egocentric vision, MM-OR can shape the future of surgical modeling by facilitating research into robust, interpretable, and dynamic OR systems. 
Complemented by MM2SG, the first multimodal large vision-language model for scene graph generation, our work establishes a new benchmark for holistic OR understanding, laying a foundation for advancing multimodal scene analysis in high-stakes environments.

\section{Acknowledgements.} The authors have been partially supported by Bundesministerium für Bildung und Forschung (BMBF) with grant [ZN 01IS17049] and by Stryker. We thank INM and Frieder Pankratz for aiding in the acquisition environment setup, and to the entire clinical team, including Prof. Dr. Michael Nogler, David Putzer, Prof. Dr. Florian Pohlig, and Prof. Dr. Rüdiger von Eisenhart-Rothe, whose contributions were vital to MM-OR.

\label{sec:conclusion}

%% file: sec/X_suppl.tex
\clearpage
\setcounter{page}{1}
\maketitlesupplementary

\begin{table}[ht]
    \centering
    \begin{minipage}[t]{0.48\linewidth}
        \vspace{0pt} 
        \centering
        \begin{tabular}{lr}
            \toprule
            \textbf{Entity} & \textbf{Count} \\
            \midrule
            Anaesthetist & 14,853 \\
            Anesthesia Eq. & 4,891 \\
            Assistant Surg. & 25,831 \\
            C-Arm & 731 \\
            Circulator & 12,225 \\
            Drape & 31,525 \\
            Drill & 2,005 \\
            Hammer & 401 \\
            Head Surgeon & 27,583 \\
            Instrument & 17,544 \\
            Instr. Table & 32,775 \\
            Mako Robot & 14,062 \\
            Monitor & 738 \\
            MPS & 25,895 \\
            MPS Station & 14,411 \\
            Nurse & 39,397 \\
            Op. Table & 30,266 \\
            Patient & 73,671 \\
            Saw & 2,874 \\
            Student & 2,432 \\
            Tracker & 877 \\
            \bottomrule
        \end{tabular}
    \end{minipage}%
    \hfill
    \begin{minipage}[t]{0.48\linewidth}
        \vspace{0pt} 
        \centering
        \begin{tabular}{lr}
            \toprule
            \textbf{Predicate} & \textbf{Count} \\
            \midrule
            Assisting & 4,635 \\
            Calibrating & 1,721 \\
            Cementing & 48 \\
            Cleaning & 113 \\
            CloseTo & 67,148 \\
            Cutting & 123 \\
            Drilling & 1,539 \\
            Hammering & 269 \\
            Holding & 23,487 \\
            Lying On & 45,924 \\
            Manipulating & 14,273 \\
            Preparing & 11,681 \\
            Sawing & 2,383 \\
            Scanning & 69 \\
            Suturing & 132 \\
            Touching & 13,963 \\
            \bottomrule
        \end{tabular}
    \end{minipage}
        \caption{Entity and predicate counts in the scene graph annotations of the MM-OR Dataset.}
            \label{tab:entity_predicate_counts}

\end{table}

\begin{table}[ht]
    \centering
    \begin{tabular}{lrrr}
        \toprule
        \textbf{Split} & \textbf{Timepoints} & \textbf{Annotations} \\
        \midrule
        Train & 37,612 & 11,123 \\
        Validation & 11,053 & 4,880\\
        Test & 13,606 & 2,960  \\
        Short Clips & 4,725 & 290\\
        \bottomrule
    \end{tabular}
    \caption{Statistics across dataset splits, timepoint and annotations.}
    \label{tab:split_stats}
\end{table}

\section{Dataset Statistics}
\label{sec:dataset_statistics}
The MM-OR dataset consists of 92,983 timepoints, with 25,277 of them annotated with panoptic segmentations and scene graphs. This total is derived from 17 full-length videos, each approximately 90 minutes long, and 22 shorter clips ranging from 1 to 10 minutes each. To ensure consistency across modalities, all data streams were synchronized to a uniform rate of 1 frame per second (FPS). Table~\ref{tab:entity_predicate_counts} provides a detailed count of annotated entities and predicates. The dataset is divided into training, validation, test splits, and short clips summarized in Table~\ref{tab:split_stats}. 

\section{Annotation Methodology}
\label{sec:annotation_methodology}
The MM-OR dataset includes over 25,000 manually annotated segmentations and scene graphs, a process that required significant effort and custom tooling. All scene graph labels were drawn from a fixed set, curated in collaboration with practicing surgeons to ensure clinical relevance. Each annotation was performed by one annotator and independently reviewed by a second annotator. On average, each scene graph comprises 8 nodes and 10 edges, with a maximum of 16 nodes and 20 edges.

\section{Detailed Scene Graph Results}
\label{sec:detailed_results}
\begin{table}[ht]
\centering
\setlength{\tabcolsep}{8pt} 
\begin{tabular}{lccc}
\toprule
\textbf{Predicate} & \textbf{Precision} & \textbf{Recall} & \textbf{F1-Score} \\
\midrule
Assisting        & 0.421 & 0.263 & 0.324 \\
Calibrating      & 0.962 & 0.212 & 0.347 \\
Cleaning         & 0.333 & 0.667 & 0.444 \\
Close to          & 0.803 & 0.636 & 0.710 \\
Cutting          & 0.000 & 0.000 & 0.000 \\
Drilling         & 0.861 & 0.369 & 0.516 \\
Hammering        & 0.708 & 0.548 & 0.618 \\
Holding          & 0.792 & 0.409 & 0.539 \\
Lying on          & 0.856 & 0.750 & 0.799 \\
Manipulating     & 0.760 & 0.699 & 0.728 \\
Preparing        & 0.699 & 0.845 & 0.765 \\
Sawing           & 0.927 & 0.722 & 0.812 \\
Scanning         & 0.500 & 0.167 & 0.250 \\
Suturing         & 0.000 & 0.000 & 0.000 \\
Touching         & 0.615 & 0.643 & 0.629 \\
\midrule
\textbf{Macro Avg} & 0.638 & 0.495 & 0.529 \\
\textbf{Weighted Avg} & 0.792 & 0.642 & 0.703 \\
\bottomrule
\end{tabular}
\caption{Per-predicate performance of MM2SG on the MM-OR dataset. Common predicates such as \textit{close to} and \textit{lying on} achieve strong performance, while rare predicates show lower scores due to limited training samples.}
\label{tab:per_predicate_results}
\end{table}

\begin{table}[ht]
\centering
\small
\begin{tabular}{lcccc}
\toprule
\textbf{Drop Chance} & 0\% & 25\% & 50\% & 75\% \\
\midrule
\textbf{F1} & 0.671 & 0.728 & \textbf{0.733} & 0.718 \\
\bottomrule
\end{tabular}
\caption{Validation results for MM2SG with varying modality drop chances. A 50\% drop chance yields the highest F1-score, indicating optimal robustness.}
\label{tab:modality_dropping}
\end{table}

\begin{table}[ht]
\centering
\small
\begin{tabular}{lccc}
\toprule
\textbf{Method} & \textbf{Head} & \textbf{Body} & \textbf{Tail} \\
\midrule
PSG~\cite{yang2022panoptic} & 0.473  & 0.156   & 0.052  \\
ORacle~\cite{ozsoy2024oracle} & 0.690  & 0.456   & 0.120  \\
MM2SG  & \textbf{0.695}  & \textbf{0.500}  & \textbf{0.262}  \\
\bottomrule
\end{tabular}
\caption{Performance of PSG, ORacle, and MM2SG across predicate frequency groups on the MM-OR dataset. MM2SG excels, particularly on rare (tail) predicates.}
\label{tab:freq_results}
\end{table}

\begin{figure*}[t]
  \centering  
   \includegraphics[width=1.0\linewidth]{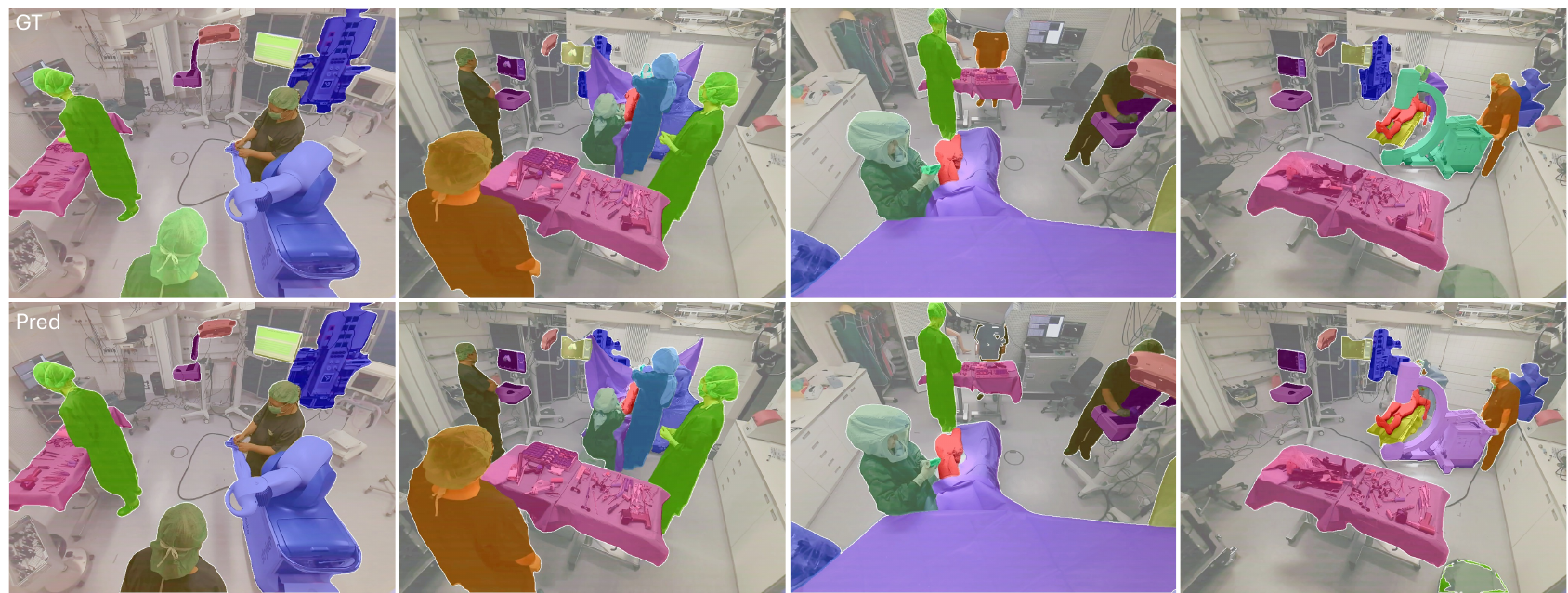}
   \caption{Qualitative segmentation results examples from a test take in MM-OR. The top row shows the ground truth segmentations and the bottom row shows the corresponding predictions.}
   \label{fig:qual_seg_results}
\end{figure*}

In Table~\ref{tab:per_predicate_results}, we provide more detailed results for our MM2SG model, where we report the results on each predicate. While most high to mid frequency classes can be predicted rather well the very rare predicates, such as \textit{cutting} and \textit{suturing} are very challenging. Our chosen macro-averaged metric emphasizes these errors by giving equal weight to all classes, highlighting the model's struggles with rare predicates, whereas weighted averaging would downplay these issues by being dominated by the performance on frequent classes. We further assess the effect of varying the modality dropping chance, where modalities are randomly omitted with a given probability. Table~\ref{tab:modality_dropping} reports F1-scores on the validation set for drop chances of 0\%, 25\%, 50\%, and 75\%. The peak performance at 50\% (F1 = 0.733) suggests that moderate modality dropping enhances generalization by reducing over-reliance on any single modality, while higher dropping (75\%) slightly degrades performance. Finally, to analyze performance across predicate frequencies, we compared MM2SG against PSG~\cite{yang2022panoptic} and ORacle~\cite{ozsoy2024oracle} by grouping predicates into head ($\geq$10,000 occurrences, e.g., \textit{lying on}), body (1,000–10,000, e.g., \textit{sawing}), and tail ($<$1,000, e.g., \textit{hammering}) categories. Table~\ref{tab:freq_results} shows that MM2SG consistently outperforms both baselines, with the largest relative gains on tail predicates (F1 = 0.262 vs. ORacle’s 0.120). This underscores MM2SG’s strength in handling rare relationships, likely due to the diverse and extensive training data in MM-OR.

\section{Qualitative Segmentation Results}
\label{sec:more_seg_results}
In Figure~\ref{fig:qual_seg_results}, we present qualitative examples of panoptic segmentations from multiple views. Results show that the baseline segmentation model performs quite well, however some misclassifications remain. These emphasize the potential for further improvement in complex scenarios, especially in regard to multiview and temporally consistent segmentations.

\section{Overview of the Surgical Procedure}
\label{sec:surgerical_procedure}
Robotic-assisted knee replacement, is a highly precise procedure designed to improve patient outcomes by optimizing implant alignment and joint function. This surgery involves replacing damaged cartilage and bone in the knee joint with artificial components to alleviate pain and restore mobility, commonly addressing conditions such as osteoarthritis.
The procedure follows roughly the following workflow:  

\begin{enumerate}
    \item \textbf{Preoperative CT Scan and Planning} \\
    A preoperative CT scan of the patient’s knee is used to create a 3D model of the joint. This enables the surgeon to develop a detailed surgical plan, including optimal alignment, sizing, and placement of the implants.
    \item \textbf{Preparation in the Operating Room (OR)} \\
    The surgical team prepares the instruments and calibrates the robotic system. The robot technician performs operational checks and calibration, while the scrub nurse and technician drape the robot to maintain sterility. The patient is brought in, positioned supine, and the surgical site is cleaned and sterilized.

    \item \textbf{Tracking Array Placement and Registration} \\
    Optical tracking arrays are attached to the patient’s femur and tibia, enabling real-time tracking of their anatomy. The robotic system aligns the preoperative plan with the patient’s knee by registering key anatomical landmarks, ensuring precise guidance during surgery.

    \item \textbf{Bone Preparation and Implant Placement} \\
    Guided by the robotic arm, the surgeon makes precise bone cuts, assisted by haptic feedback that restricts movement to predefined boundaries. This minimizes tissue damage and ensures accurate preparation for the implants. The implants are placed according to the surgical plan, with some intraoperative adjustments.
    
    \item \textbf{Closure and Postoperative Verification} \\
    After implant placement, the surgical site is cleaned, and the wound is closed. The tracking arrays are removed, and the robotic system is shut down. To confirm proper alignment of the implants, an intraoperative X-ray scan is performed before the patient leaves the OR.
\end{enumerate}

This workflow highlights the precision and integration of robotic assistance in modern knee replacement surgeries. The MM-OR dataset captures these steps in detail, providing a comprehensive resource for studying robotic-assisted surgical workflows.

\section{Technical Setup}
\label{sec:tech_setup}
The MM-OR dataset was acquired using a multimodal recording setup designed to capture the complex dynamics of robotic knee replacement surgeries. We used multiple ceiling-mounted Azure Kinect cameras for RGB-D recordings, AXIS Q6125-LE PTZ Network Cameras for detailed views, and Sennheiser SK 300 G4-RC wireless microphone systems for the audio recordings, worn by the head surgeon, assistant surgeon and robot technician. Tracking data and robot logs were directly extracted from the robotic surgery setup, and the robot interface was recorded using an HDMI splitter. To maintain synchronization across all modalities at 1 FPS, high-resolution streams were downsampled.

\section{Specimen Preparation}

\begin{figure}
\centering
\includegraphics[width=.75\linewidth]{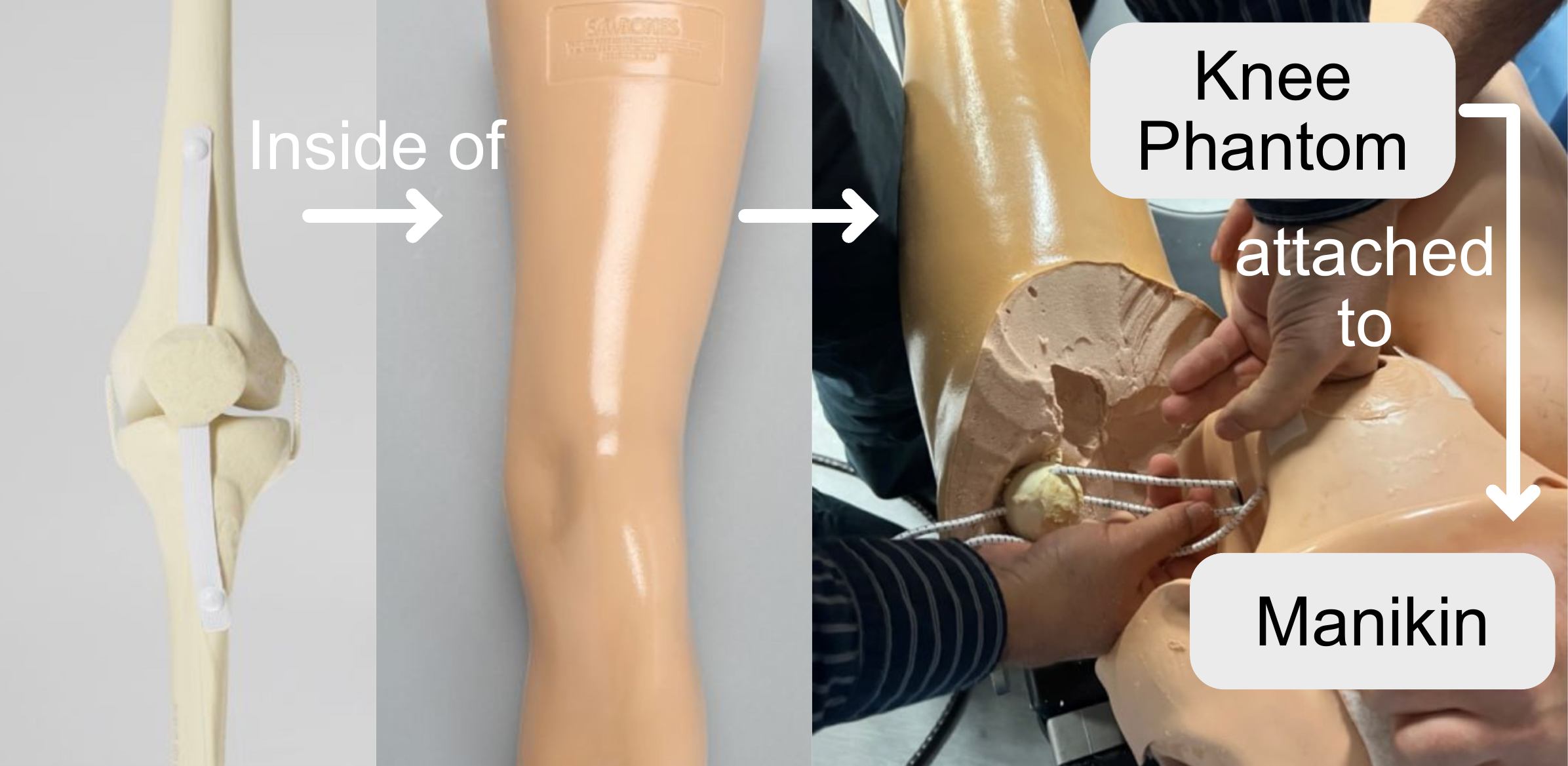}
\caption{Prepared specimen: Bone phantom with ligaments (left), enclosed in foam for tissue simulation (middle), attached to manikin (right)}
\label{fig:phantom}
\end{figure}

To ensure high realism, we used professional-grade knee phantoms from a commercial supplier~\footnote{https://www.sawbones.com}. We performed realistic bone cuts and implant placements, using a new phantom for each acquisition. Fig. \ref{fig:phantom} shows the prepared phantom setup.

\section{Supplementary Video}
\label{sec:supp_video}

To provide a summary of the MM-OR dataset, we include a supplementary video. This video overlays all recorded modalities onto a dynamic 3D point cloud, offering a comprehensive visualization of the surgical environment and multimodal data. While some faces are visible in the video, no individual recognizable in the footage has any association with our institution or the authors.